\documentclass{article}
\usepackage{iclr2027_conference,times}
\usepackage[T1]{fontenc}
\usepackage{amsmath,amssymb,booktabs,graphicx,hyperref,url}
\usepackage[ruled,vlined,linesnumbered]{algorithm2e}
\SetAlgoNlRelativeSize{0}
\title{COVER: Learning to Accept More\\in Selective Sleep Staging}
\author{%
\begin{minipage}[t]{0.97\textwidth}
\centering\normalsize
Yukai Song\textsuperscript{1}\quad
Yangfan Deng\textsuperscript{2}\quad
Jijun Yin\textsuperscript{1}\quad
Zhi-Hong Mao\textsuperscript{1}\quad
Jingtong Hu\textsuperscript{1}\\[0.7ex]
\normalfont\small
\textsuperscript{1}Department of Electrical and Computer Engineering,\\
University of Pittsburgh, Pittsburgh, PA, USA\\[0.4ex]
\textsuperscript{2}Department of Electrical and Computer Engineering,\\
University of Maryland, College Park, MD, USA
\end{minipage}%
}
\hypersetup{pdfauthor={Yukai Song, Yangfan Deng, Jijun Yin, Zhi-Hong Mao, Jingtong Hu}}

\newcommand{\ind}{\mathbf{1}}
\newcommand{\RSE}{R_{\mathrm{SE}}}
\newcommand{\Rpool}{R_{\mathrm{pool}}}

\newcommand{\exHistFixedScaleCoverage}{50.40}
\newcommand{\exHistFixedScaleRisk}{4.53}

\newcommand{\rvNestedFixedBceCovAbs}{1.65}
\newcommand{\rvNestedFixedBceCovOne}{1.7}

\newcommand{\rvNestedFixedBceCovHigher}{18}

\newcommand{\rvHistFixedSeleCovAbs}{1.05}

\newcommand{\rvHistFixedSeleCovHigher}{9}

\newcommand{\rvHistFixedDuofCovAbs}{6.78}

\newcommand{\rvHistFixedDuofCovHigher}{14}

\newcommand{\rvHistFixedCovOne}{50.4}
\newcommand{\rvHistFixedRiskOne}{4.5}
\newcommand{\rvHistSeleCovOne}{49.3}

\newcommand{\rvMspCovOne}{31.5}

\newcommand{\rvDuofCovOne}{43.6}

\newcommand{\rvFixedNOneB}{831}

\newcommand{\rvFixedNOneQ}{19}

\newcommand{\rvFixedNThreeAccuracy}{96.99}

\newcommand{\rvFixedErrors}{982}

\newcommand{\rvFixedTransitionErrors}{525}
\newcommand{\rvExchangeFixedOnlyAccepted}{2,321}
\newcommand{\rvExchangeFixedOnlyErrors}{244}
\newcommand{\rvExchangeBceOnlyAccepted}{1,505}
\newcommand{\rvExchangeBceOnlyErrors}{129}
\newcommand{\rvExchangeCommonAccepted}{19,134}
\newcommand{\rvExchangeCommonErrors}{738}

\newcommand{\lbMatchedBudget}{20,639}
\newcommand{\lbMspCovCI}{+17.34 [+9.74, +25.54]}
\newcommand{\lbMspRiskCI}{+0.21 [-1.29, +1.81]}
\newcommand{\lbMspMatchedErrors}{1,467}
\newcommand{\lbLinearMatchedErrors}{867}

\newcommand{\lbMspFewer}{16}
\newcommand{\lbMspPoolCI}{-2.91 [-5.37, -1.09]}
\newcommand{\lbFklMatchedErrors}{2,284}
\newcommand{\lbPrimaryMatchedErrors}{1,198}
\newcommand{\lbPrimaryPoolCI}{-1.60 [-3.07, -0.63]}
\newcommand{\lbPrimaryCovCI}{+12.55 [+6.81, +19.25]}
\newcommand{\lbDuofMatchedErrors}{1,031}
\newcommand{\lbDuofPoolCI}{-0.79 [-2.08, -0.02]}
\newcommand{\lbDuofCovCI}{+5.17 [+0.98, +9.83]}
\newcommand{\lbDuofFewerTotal}{164}
\newcommand{\lbDuofSubjectFourteen}{107}
\newcommand{\lbDuofDeleteRange}{[-0.95, -0.29]}
\newcommand{\lbOriginalMatchedErrors}{912}
\newcommand{\lbOriginalMinusLinearPoolCI}{+0.22 [-0.18, +0.72]}
\newcommand{\lbOriginalMinusLinearCovCI}{-1.72 [-6.32, +1.36]}

\newcommand{\lbMspCovDelta}{17.34}

\newcommand{\lbFixedMinusLinearPoolAbs}{0.11}

\newcommand{\lbExchangeCommonRisk}{3.9}
\newcommand{\lbExchangeFixedOnlyRisk}{10.5}
\newcommand{\lbExchangeBceOnlyRisk}{8.6}
\newcommand{\lbLinearCovOne}{48.8}

\newcommand{\lbCurveLinearSpread}{0.41}
\newcommand{\lbFixedMinusLinearPoolCI}{+0.11 [-0.18, +0.48]}

\iclrfinalcopy
\hypersetup{pdftitle={COVER: Learning to Accept More in Selective Sleep Staging}}
\begin{document}
\maketitle
\lhead{Preprint}
\begin{abstract}
Traditional sleep-staging methods primarily pursue higher accuracy through increasingly capable models, applying the same model to every EEG epoch regardless of difficulty. Such uniform deployment expends computation on epochs that a smaller model could handle reliably, motivating cascades in which a primary classifier accepts its reliable predictions and defers the remainder to a more capable model. In this paper, we study the first stage of such a cascade: maximizing the coverage of fixed primary predictions subject to a prescribed accepted-risk target. We propose COVER (COVerage-oriented Error Ranking), which integrates two key innovations: (i)~auxiliary-informed primary-error learning, which replaces maximum softmax probability (MSP) with a learned error score while preserving the primary labels, and (ii)~fixed-scale scorer refinement, which builds on this score to directly maximize coverage under an empirical accepted-risk constraint rather than error-prediction accuracy over all epochs. We evaluate COVER on Sleep-EDF-20 at a 5\% accepted-risk target, with subjects held out from all fitting and selection. Auxiliary-informed error learning raises mean subject coverage from \rvMspCovOne\% for MSP to \lbLinearCovOne\% at similar subject-equal risk. At equal acceptance volume, with MSP accepting the same number of epochs as the learned scorer in each subject (\lbMatchedBudget{} in total), errors fall from \lbMspMatchedErrors{} to \lbLinearMatchedErrors{}. Fixed-scale refinement then adds \rvNestedFixedBceCovOne{} percentage points of coverage over its initialization in nested development, and COVER attains the highest mean coverage among eight evaluated scorers, \rvHistFixedCovOne\% at \rvHistFixedRiskOne\% subject-equal risk, above the selective-ranking baseline SELE (\rvHistSeleCovOne\%) and the probability-fusion comparator DuoF (\rvDuofCovOne\%). To the best of our knowledge, this is the first work to combine auxiliary-informed primary-error learning with fixed-scale coverage refinement for selective sleep staging, offering a basis for reliability-aware allocation of computation in cascaded sleep staging.
\end{abstract}

\section{Introduction}
Sleep staging assigns each 30-second epoch of an overnight recording to a sleep stage. The resulting hypnogram summarizes sleep architecture and supports assessment of sleep disorders \citep{fiorillo2019automated}. Manual scoring is labor-intensive, with approximately 83\% inter-scorer agreement and particular difficulty in N1 \citep{rosenberg2013aasm}. Deep models learn EEG representations and temporal context \citep{supratak2017deepsleepnet,phan2019seqsleepnet,perslev2019utime,eldele2021attention,phan2022xsleepnet} and support clinical analysis \citep{stephansen2018neural}. Yet high aggregate accuracy does not make every prediction reliable: transitional ambiguity, stage imbalance, and inter-subject variability leave some predictions wrong, including confident ones \citep{hong2021confidence,lin2022scrib}.

Applying a more capable model to every epoch also spends computation on predictions a smaller model already gets right. Adaptive inference instead allocates computation or expertise through cascades, early exits, and deferral \citep{viola2001rapid,wang2018idk,teerapittayanon2016branchynet,huang2018multiscale,bolukbasi2017adaptive,madras2018predict,mozannar2020consistent}. A sleep-staging cascade can therefore accept reliable primary predictions and defer the remainder. Its first stage is a selective classification problem \citep{chow1970optimum,elyaniv2010foundations,geifman2017selective}: maximize \emph{coverage}, the fraction accepted, at a prescribed \emph{accepted risk}, the error rate among accepted predictions. A useful first stage must identify as many reliable predictions as possible without spending the error budget on confidently wrong answers. We study this problem for a fixed primary classifier; a complete second stage remains future work.

The first challenge is the acceptance signal. Maximum softmax probability (MSP) is a standard baseline \citep{hendrycks2017baseline,geifman2017selective}, but average calibration need not improve ranking \citep{guo2017calibration}. Equal MSP values can conceal different runner-up stages, distribution shapes, or auxiliary-model agreement. These differences may carry information about primary errors that a single confidence value discards. Prior work estimates reliability from outputs or representations \citep{jiang2018trust,corbiere2019addressing,kamath2020selective}, including fused predictions from auxiliary models \citep{zhou2025asymmetric}. COVER learns primary errors from both models' outputs, using an auxiliary model with a different input representation. Disagreement is interpreted alongside confidence because either model may be wrong; accepted labels remain the primary predictions.

The second challenge is the training objective. Binary cross-entropy (BCE) rewards error prediction across all epochs, whereas first-stage utility depends only on accepted epochs: a good average error predictor need not yield the largest low-risk acceptance set. Joint selective training changes the classifier \citep{geifman2019selectivenet,liu2019deep}; risk-control methods select thresholds for a given score \citep{geifman2017selective,bates2021distribution,angelopoulos2025learn}; SELE learns a given classifier's uncertainty by minimizing a convex surrogate of the area under the risk--coverage curve \citep{franc2019uncertainty}. A scorer with limited capacity may need to prioritize distinctions within the low-risk region over probability accuracy elsewhere. We therefore refine the error score toward coverage at the prescribed risk target while keeping the primary classifier fixed.

\textbf{COVER (COVerage-oriented Error Ranking)} first learns primary errors from primary and auxiliary outputs, then applies \textbf{FixedScale} refinement under an empirical risk constraint at a fixed reference score scale. Figure~\ref{fig:method} and Algorithm~\ref{alg:cover} summarize this progression.

Our contributions are:
\begin{enumerate}
\item \textbf{A coverage-oriented first-stage framework.} COVER combines auxiliary-informed primary-error learning with fixed-scale coverage refinement, preserving every accepted primary label.
\item \textbf{Auxiliary information recovers useful predictions.} On Sleep-EDF-20, at equal acceptance volume, a linear error scorer reduces errors from \lbMspMatchedErrors{} for MSP to \lbLinearMatchedErrors{}, compared with \lbPrimaryMatchedErrors{} for primary-only learning even though the latter selects the best of eight scorers (Table~\ref{tab:operating}).
\item \textbf{Refinement improves the operating point.} FixedScale adds \rvNestedFixedBceCovOne{} percentage points (pp) of coverage over its BCE initialization in nested development within the same cohort. It has the highest observed mean coverage among eight scorers in the main comparison (Figure~\ref{fig:operating}).
\end{enumerate}

\section{Related Work}
\paragraph{Selective classification and error scoring.}
The reject option \citep{chow1970optimum} trades coverage against accepted risk \citep{elyaniv2010foundations,geifman2017selective}. Alternatives to MSP \citep{hendrycks2017baseline} include neighborhood-based trust \citep{jiang2018trust}, true-class probability regression \citep{corbiere2019addressing}, and model-behavior calibrators \citep{kamath2020selective}. These estimators seek informative acceptance scores without relying solely on the largest class probability. SelectiveNet and Deep Gamblers jointly train classification and rejection \citep{geifman2019selectivenet,liu2019deep}, although their gains stem mainly from a better-trained classifier rather than from the selection mechanism \citep{feng2023towards}. SELE instead learns uncertainty for a given classifier by minimizing a convex surrogate of the area under the risk--coverage curve \citep{franc2019uncertainty}. COVER shares this setting but refines a learned primary-error score toward coverage at a single risk target using a soft constrained objective, rather than optimizing overall ordering. SELE is therefore a direct comparator with the same features and linear capacity.

\paragraph{Auxiliary evidence and probability fusion.}
Asymmetric Duos learn weighted logit fusion with a smaller sidekick; their \emph{UQ Only} setting preserves the primary label \citep{zhou2025asymmetric}. DuoF adapts this setting to COVER's two component models (Appendix~\ref{app:duof}). COVER instead supervises primary error using output summaries and disagreement. Thus the distinction is the supervised target and use of the available evidence, not whether primary labels are preserved. Feature-based sleep staging provides EEG representations distinct from end-to-end features \citep{mao2022finding,deng2024umap,deng2024twostage}; COVER uses such a model as auxiliary evidence while retaining primary labels.

\paragraph{Sleep-staging reliability and adaptive computation.}
Learned confidence flags sleep-stage predictions for review \citep{hong2021confidence}, while SCRIB constructs set-valued predictions under class-specific risk constraints \citep{lin2022scrib}. Cascades, early exits, and model selection allocate computation by difficulty \citep{viola2001rapid,wang2018idk,teerapittayanon2016branchynet,huang2018multiscale,bolukbasi2017adaptive}; learning to defer routes inputs to another decision maker \citep{madras2018predict,mozannar2020consistent}. COVER studies first-stage acceptance for sleep staging, determining which unchanged primary predictions can be retained at a risk target and which epochs remain for subsequent processing.

\paragraph{Calibration and risk control.}
Probability calibration aligns probabilities with outcome frequencies \citep{guo2017calibration}; \citet{fisch2022calibrated} learn a selector for calibration within the accepted subset. Because a strictly increasing recalibration preserves a score's ordering, it leaves the acceptance sets attainable by thresholding unchanged; selective performance therefore depends on ordering rather than calibration. Distribution-free methods provide finite-sample risk guarantees under stated assumptions \citep{bates2021distribution,angelopoulos2025learn,bai2026conformal}, complementing informative score learning. COVER uses empirical rule selection without such a guarantee. Because rejection can distribute benefits unevenly \citep{jones2021selective}, we report stage-wise retention (Section~\ref{sec:resultsstages}) and subject-level results (Appendix~\ref{app:revision}).

\section{Problem Setup}
\label{sec:setup}
\subsection{Fixed-label selective prediction}
Subject $s$ contributes $n_s>0$ EEG epochs $x_{si}$ with true stages $y_{si}\in\mathcal Y=\{\mathrm{W},\mathrm{N1},\mathrm{N2},\mathrm{N3},\mathrm{REM}\}$. The primary (target) classifier outputs $p_T(j\mid x)$ from raw EEG and predicts $\hat y_T(x)=\arg\max_j p_T(j\mid x)$; an auxiliary (feature-based) classifier outputs $p_F(j\mid x)$ from handcrafted features of the same epoch. Both remain fixed during acceptance-score learning and inference.

Define primary error $e_{si}=\ind\{\hat y_T(x_{si})\ne y_{si}\}$. Rule $\pi$ assigns acceptance $a_\pi(x)\in\{0,1\}$ from a score $r(x)$, where larger values indicate greater error risk. It accepts when $r(x)<\tau$, with label-free handling of boundary ties (Appendix~\ref{app:acceptance}). Acceptance releases $\hat y_T(x)$; rejection withholds it for later processing. Neither the auxiliary model nor the score changes this label. Ground truth supplies fitting targets and evaluation outcomes, never inference-time features.

\subsection{Subject-weighted coverage, risk, and objective}
To prevent recording length from determining influence, assign $w_{si}=1/(Sn_s)$ over $S$ subjects. Each subject has total weight $1/S$. With $a_{si}=a_\pi(x_{si})$, accepted counts $A_s=\sum_i a_{si}$, and accepted errors $K_s=\sum_i a_{si}e_{si}$, define
\begin{equation}
\begin{aligned}
C&=\sum_{s,i}w_{si}a_{si}=\frac1S\sum_s\frac{A_s}{n_s},\\
\RSE&=\frac{\sum_{s,i}w_{si}a_{si}e_{si}}{\sum_{s,i}w_{si}a_{si}}
=\frac{\sum_s K_s/n_s}{\sum_s A_s/n_s},\qquad
\Rpool=\frac{\sum_s K_s}{\sum_s A_s}.
\end{aligned}
\label{eq:metrics}
\end{equation}
$C$ is mean subject coverage; $1-C$ is the mean fraction deferred. $\RSE$ weights individual accepted risks by subject coverage, rather than averaging them equally; $\Rpool$ pools accepted epochs. They coincide when recording lengths are equal. Zero-acceptance subjects remain in $C$ and contribute zero error and acceptance mass; both risks are undefined under all-abstention.
At prespecified $\alpha\in(0,1)$, a subject-weighted form of the selective-classification objective \citep{geifman2017selective} is
\begin{equation}
\max_{\substack{a\in\mathcal A\\C(a)>0}} C(a)
\quad\text{s.t.}\quad \RSE(a)\le\alpha,
\label{eq:task}
\end{equation}
where $\mathcal A$ preserves the primary labels. Section~\ref{sec:method} pursues this objective through empirical score learning and rule selection.

\section{Method}
\label{sec:method}
\begin{figure}[t]
\centering\includegraphics[width=\linewidth]{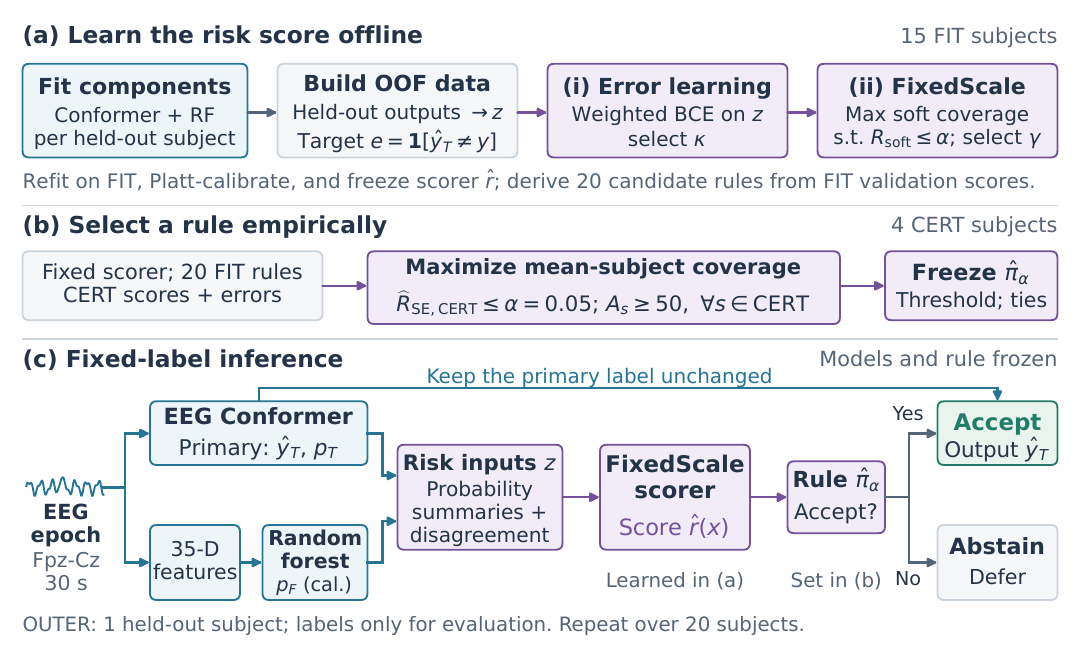}
\caption{Overview of COVER. (a)~FIT component out-of-fold (OOF) outputs supply risk vectors $z$ and primary-error targets $e$. Step~(i) fits a subject-weighted BCE error score; step~(ii) refines it toward coverage at the risk target; the refitted scorer is then Platt-calibrated and frozen. (b)~CERT selects an empirical acceptance rule with the scorer frozen. (c)~At inference, the frozen scorer and rule determine acceptance; the blue bypass releases the primary label unchanged. The waveform is schematic.}
\label{fig:method}
\end{figure}

COVER builds the acceptance score in two steps (Figure~\ref{fig:method}, Algorithm~\ref{alg:cover}). Step~(i) learns where the primary classifier errs from the outputs of both models, replacing MSP with a learned error score (Section~\ref{sec:errorlearning}). Step~(ii) refines this score toward coverage under an empirical accepted-risk constraint (Section~\ref{sec:refinement}). In both steps the released label remains $\hat y_T$; Section~\ref{sec:protocol} describes configuration selection and rule locking on separate subjects. We instantiate the primary classifier as an EEG Conformer \citep{song2023eeg} and the auxiliary model as a random forest (RF) on handcrafted features.

\subsection{Step (i): auxiliary-informed primary-error learning}
\label{sec:errorlearning}
The risk vector $z\in\mathbb R^{37}$ describes what the two models say about an epoch. Fourteen primary features summarize $p_T$: the five class probabilities, five predicted-class indicators, MSP, the runner-up probability, the top-two margin, and normalized entropy; fourteen analogous features summarize $p_F$. Nine disagreement features compare the two distributions, including forward, reverse, and symmetric KL divergence, Jensen--Shannon divergence, label disagreement, and the probability each model assigns to the other's predicted class (Appendix~\ref{app:implementation}). Disagreement is paired with both models' confidence because disagreement alone is ambiguous: it can signal either a primary or an auxiliary error. The RF's input features do not enter $z$, so COVER and DuoF draw on outputs of the same two components.

The initial error score is a logistic regression of the primary error with the subject weights of Equation~\ref{eq:metrics}:
\begin{equation}
\min_{\theta,b}\ \sum_{s,i}w_{si}\bigl[\log(1+\exp(u_{si}))-e_{si}u_{si}\bigr]
+\frac{\|\theta\|_2^2}{2N\kappa},\qquad u_{si}=z_{si}^{\top}\theta+b.
\label{eq:bce}
\end{equation}
The bracket is the binary cross-entropy (BCE) for predicting $e_{si}$, and larger $u$ indicates a more likely primary error. Features are standardized on the current risk-training subjects; $N$ counts training epochs, $\kappa>0$ is the inverse regularization strength, and the intercept is unpenalized. Ground truth enters only through the training targets, so at inference the score depends on model outputs alone.

\subsection{Step (ii): fixed-scale coverage refinement}
\label{sec:refinement}
BCE rewards error prediction on every epoch, but the task in Equation~\ref{eq:task} is decided only by the epochs that are accepted. FixedScale closes this gap by optimizing a smooth version of the task itself. Replacing the hard decision $a_{si}$ with a sigmoid gate gives differentiable coverage and accepted risk:
\begin{equation}
g_{si}=\sigma\!\left(\frac{t-z_{si}^{\top}\beta}{T_0}\right),\quad
C_{\rm soft}=\sum_{s,i}w_{si}g_{si},\quad
R_{\rm soft}=\frac{\sum_{s,i}w_{si}e_{si}g_{si}}{C_{\rm soft}},
\label{eq:soft}
\end{equation}
where $t$ is a training threshold and $T_0>0$ a fixed temperature; replacing $g_{si}$ by $\ind\{z_{si}^{\top}\beta<t\}$ recovers $C$ and $\RSE$ for a nonempty acceptance set. Refinement solves the smooth, regularized counterpart of Equation~\ref{eq:task},
\begin{equation}
\max_{\beta,t}\ C_{\rm soft}-\frac{\gamma}{2}\|\beta\|_2^2
\quad\text{s.t.}\quad R_{\rm soft}\le\alpha,\quad C_{\rm soft}\ge10^{-4},
\label{eq:refine}
\end{equation}
where the lower bound keeps the risk denominator away from zero. Features, subject weights, and linear capacity are those of step~(i); only the objective changes.

\paragraph{Why the objective matters.}
A linear scorer cannot rank every epoch correctly, so the two objectives can prefer different parameters. BCE's gradient weights each epoch by its residual $\sigma(u_{si})-e_{si}$, wherever the epoch lies. In refinement, each epoch's contribution to the coverage and risk-constraint gradients is scaled by $g_{si}(1-g_{si})/T_0$, which peaks at the soft acceptance boundary and is small in saturated regions. Refinement therefore spends linear capacity on epochs whose acceptance is in question, rather than on epochs that are clearly accepted or clearly rejected.

\paragraph{Fixed scale.}
The gate depends on $(\beta,t,T_0)$ only through $(t-z^{\top}\beta)/T_0$: scaling all three together leaves every gate unchanged but changes the penalty $\gamma\|\beta\|_2^2/2$, so temperature and regularization are confounded. FixedScale anchors the scale to BCE. With $\sigma_0=\max\{\operatorname{SD}_w(u^{\rm BCE}),10^{-6}\}$ and $\beta_0=\theta_{\rm BCE}/\sigma_0$, the initial score $z^{\top}\beta_0$ has unit weighted standard deviation when the floor is inactive, and $\sigma_0$, $b_{\rm BCE}$, and $T_0$ stay fixed during refinement, so $\gamma$ acts on a consistent reference scale. Because Equation~\ref{eq:refine} is nonconvex, sequential least-squares programming (SLSQP) with analytic gradients starts from prescribed BCE-based initializations and keeps the best converged feasible solution (Appendix~\ref{app:refinement}). The refined raw score is $u(x)=\sigma_0z(x)^{\top}\hat\beta+b_{\rm BCE}$; the training threshold is discarded because CERT sets the deployed rule.

\subsection{Selection and rule locking}
\label{sec:protocol}
Each outer fold assigns subjects to three roles: \emph{FIT} develops the scorer, \emph{CERT} locks the acceptance rule with the scorer frozen, and \emph{OUTER} evaluates the frozen system. Three design choices separate the data used for each decision.

\paragraph{Risk vectors from unseen subjects.}
At deployment the components score subjects they were not trained on, so the scorer should learn from such outputs. Each FIT subject's risk vector therefore comes from component OOF predictions: that subject is excluded from component training, checkpoint selection, and RF calibration. A separate component fit on designated FIT subjects scores CERT and OUTER (Appendix~\ref{app:data}).

\paragraph{Selection by the task criterion.}
Hyperparameters are chosen by the quantity the task rewards, not by BCE loss. A configuration's feasible coverage is the largest cross-validated $C$ among candidate rules satisfying $\RSE\le\alpha$, or zero if none qualifies; we select $\kappa$ for BCE, then $\gamma$ for its refinement. Validation runs over within-FIT subject folds, with each fold's scores Platt-calibrated on the other folds. Because component fits are reused across risk-level folds, and the other folds' OOF scores used for Platt fitting come from risk models that saw the current validation subject, this validation is conditional on the components (Appendix~\ref{app:implementation}). The selected scorer is then refitted on FIT, and a nondecreasing Platt map $\hat h$ fitted to its raw OOF scores gives $\hat r(x)=\hat h(u(x))$. Candidate rules $\Pi_{\rm FIT}$ are thresholds at subject-weighted quantiles $\mathcal Q$ of the calibrated validation scores, with label-free tie handling, rather than of in-sample refit scores.

\paragraph{Locking acceptance on separate subjects.}
Acceptance volume is set only after the scorer is frozen, on subjects that played no role in its development. With $\hat r$ fixed, CERT selects
\begin{equation}
\hat\pi_\alpha\in\arg\max_{\pi\in\Pi_{\rm FIT}}\widehat C_{\rm CERT}(\pi)
\quad\text{s.t.}\quad \widehat R_{\rm SE,CERT}(\pi)\le\alpha,\quad A_s(\pi)\ge m\ \forall s\in{\rm CERT}.
\label{eq:objective}
\end{equation}
The support constraint ensures that every CERT subject contributes accepted epochs; it is not a per-subject risk bound. If no candidate is feasible, COVER abstains on all epochs. OUTER then evaluates the frozen $\hat r$ and $\hat\pi_\alpha$. This is empirical rule selection, without a finite-sample guarantee (Appendix~\ref{app:acceptance}).

\begin{algorithm}[t]
\small
\KwData{FIT pairs $\{(z_{si},e_{si})\}$; CERT outputs and labels; $\alpha$, grids, rule quantiles $\mathcal Q$, support $m$}
\KwResult{Frozen scorer $\hat r$ and rule $\hat\pi_\alpha$; released labels remain $\hat y_T$}
\tcp{(i) Auxiliary-informed primary-error learning}
Select $\hat\kappa$ by feasible coverage; refit BCE (Equation~\ref{eq:bce}) on FIT to obtain $(\theta_{\rm BCE},b_{\rm BCE})$\;
\tcp{(ii) Fixed-scale coverage refinement}
$\sigma_0\gets\max\{\operatorname{SD}_w(u^{\rm BCE}),10^{-6}\}$; $\beta_0\gets\theta_{\rm BCE}/\sigma_0$\;
Select $\hat\gamma$ by feasible coverage; solve Equation~\ref{eq:refine} from BCE-based starts, keeping the best converged feasible $\hat\beta$\;
\tcp{Calibrate and build candidates}
$\hat r(x)\gets\hat h(\sigma_0z(x)^{\top}\hat\beta+b_{\rm BCE})$, with $\hat h$ fitted to raw OOF scores\;
$\Pi_{\rm FIT}\gets$ quantile rules from calibrated validation scores\;
\tcp{Lock acceptance on CERT}
Solve Equation~\ref{eq:objective} for $\hat\pi_\alpha$, or return all-abstention\;
\Return{$(\hat r,\hat\pi_\alpha)$}; release $\hat y_T(x)$ iff $a_{\hat\pi_\alpha}(x)=1$\;
\caption{COVER. FIT learns and refines the score; CERT locks acceptance.}
\label{alg:cover}
\end{algorithm}

\section{Experiments}
\label{sec:experiments}
\subsection{Setup}
We use Sleep-EDF-20: 20 subjects, 39 recordings, and 42,308 epochs \citep{sleepedf,kemp2000sleep,pollard2026physionet}. Inputs are 30-s Fpz--Cz epochs at 100\,Hz, without other channels or neighboring epochs. Each outer fold assigns 15/4/1 subjects to FIT/CERT/OUTER. All methods share primary predictions and $\alpha=0.05$ (Appendix Table~\ref{tab:settings}).

\paragraph{Evaluations.}
The main evaluation applies each CERT-locked rule to its OUTER subject. FixedScale was specified after the prespecified analysis and is evaluated on the same subjects, with $\gamma$ selected in FIT. Nested development fixes $\gamma=1$ and repeats risk-layer fitting, calibration, and rule locking inside FIT, reusing frozen component outputs (Appendix~\ref{app:upgrade}).

\paragraph{Scorers.}
Step~(i) compares MSP, forward-KL, primary-only error learning (the best of eight candidates), DuoF probability fusion \citep{zhou2025asymmetric}, and Linear-BCE on all 37 features. Step~(ii) retains this linear capacity: FixedScale refines BCE, BCEAnchor changes the penalty center to $\beta_0$, and SELE optimizes a risk--coverage surrogate \citep{franc2019uncertainty}. SELE selects its own regularization configuration.

\paragraph{Reporting.}
Own-rule results report $C$, $\RSE$, and $\Rpool$. Matched-count comparisons use Linear-BCE's accepted count in each subject (\lbMatchedBudget{} epochs), separating ranking from acceptance volume; their pooled-risk differences compare errors at equal counts. Nested records are averaged within subject before aggregation, retaining 20 evaluation units. Descriptive subject-level intervals and sensitivities are in Appendix~\ref{app:resampling}.

\subsection{Step (i): learning errors recovers useful predictions}
\label{sec:resultsinformation}
Linear-BCE increased mean subject coverage over MSP by \lbMspCovDelta{} pp at similar subject-equal risk (Table~\ref{tab:operating}). At identical per-subject acceptance counts, errors fell from \lbMspMatchedErrors{} to \lbLinearMatchedErrors{}. Learning errors therefore recovers more correct primary predictions at the same acceptance budget.

Auxiliary evidence helps when interpreted alongside model confidence. Disagreement alone incurred \lbFklMatchedErrors{} matched errors, exceeding MSP; primary-only learning reduced this to \lbPrimaryMatchedErrors{}. Combining both models' outputs improved further, despite primary-only learning selecting among eight scorers. DuoF came closest among these comparators, with \lbDuofMatchedErrors{} errors, supporting primary-error supervision over probability fusion in this cohort.

The prespecified analysis used a broader step-(i) scorer search; its results and the subsequent choice of Linear-BCE as the matching anchor are retained in Appendix~\ref{app:composition}.

\begin{table}[t]
\caption{Main evaluation of eight scorers. Coverage and risks (\%) use each own CERT-locked rule; Accepted is its epoch count. Matched errors use Linear-BCE's counts within every subject, totaling \lbMatchedBudget{} epochs per method.}
\label{tab:operating}
\centering\small\begin{tabular}{@{}lrrrrr@{}}
\toprule
Scorer & Coverage & $R_{\rm SE}$ & $R_{\rm pool}$ & Accepted & Matched errors \\
\midrule
\multicolumn{6}{@{}l}{\emph{Confidence}} \\
\quad MSP & 31.45 & 3.94 & 4.15 & 13,625 & 1,467 \\
\addlinespace[2pt]
\multicolumn{6}{@{}l}{\emph{Step-(i) components}} \\
\quad Forward-KL & 30.56 & 5.17 & 5.17 & 12,935 & 2,284 \\
\quad Primary-only & 36.24 & 3.59 & 3.67 & 15,603 & 1,198 \\
\quad DuoF & 43.62 & 4.29 & 4.36 & 18,598 & 1,031 \\
\addlinespace[2pt]
\multicolumn{6}{@{}l}{\emph{Step (i): learned error score}} \\
\quad Linear-BCE & 48.79 & 4.15 & 4.20 & 20,639 & 867 \\
\addlinespace[2pt]
\multicolumn{6}{@{}l}{\emph{Step (ii): refinement and controls}} \\
\quad \textbf{FixedScale} & 50.40 & 4.53 & 4.58 & 21,455 & 890 \\
\quad SELE & 49.35 & 4.31 & 4.35 & 20,875 & 868 \\
\quad BCEAnchor & 49.22 & 4.06 & 4.11 & 20,805 & 857 \\
\bottomrule
\end{tabular}

\end{table}

\subsection{Step (ii): refinement raises coverage beyond its initialization}
\label{sec:resultsrefinement}\label{sec:resultscoverage}
\begin{figure}[t]
\centering\includegraphics[width=\linewidth]{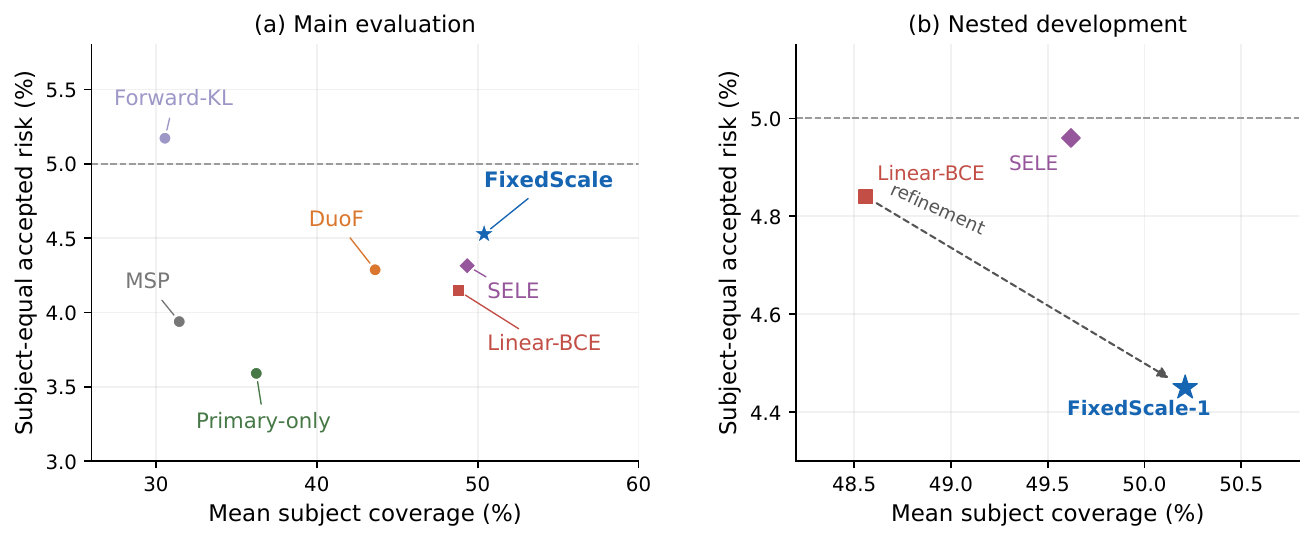}
\caption{Locked operating points. (a)~Main evaluation; Table~\ref{tab:operating} also includes BCEAnchor. (b)~Nested development within FIT; the arrow marks Linear-BCE to FixedScale-1 refinement. Axes show mean subject coverage and subject-equal accepted risk. Dashed lines mark the 5\% target; evaluation labels do not select these rules.}
\label{fig:operating}
\end{figure}

FixedScale achieved the highest observed mean coverage among eight scorers, \exHistFixedScaleCoverage\% at \exHistFixedScaleRisk\% subject-equal risk (Figure~\ref{fig:operating}a). Coverage exceeded DuoF by \rvHistFixedDuofCovAbs{} pp and SELE by \rvHistFixedSeleCovAbs{} pp; the latter difference is within sampling variation (Appendix~\ref{app:revision}).

Nested development tests the increment over BCE initialization within the same cohort. Fixed-strength refinement added \rvNestedFixedBceCovAbs{} pp of locked-rule coverage, a gain whose interval excludes zero, with a lower risk point estimate (Figure~\ref{fig:operating}b; Appendix Table~\ref{tab:nested}). Features and linear capacity remain fixed.

The evidence supports an operating-point gain without establishing uniformly better ranking. At matched counts, FixedScale and Linear-BCE differed by \lbFixedMinusLinearPoolAbs{} pp in pooled risk. Their quota curves and SELE's were also close (Appendix Figure~\ref{fig:quota}). BCEAnchor, which centers the penalty at $\beta_0$, yielded lower coverage point estimates than FixedScale (Appendix~\ref{app:upgrade}).

\subsection{What the accepted sets contain}
\label{sec:resultssearch}\label{sec:resultsstages}
Acceptance concentrates in stages the primary classifier handles well. Accepted true-Wake, N2, and N3 predictions were at least \rvFixedNThreeAccuracy\% accurate. N1 remained difficult: only \rvFixedNOneQ{} of \rvFixedNOneB{} correct true-N1 predictions were retained; N1 and REM supplied \rvFixedTransitionErrors{} of \rvFixedErrors{} accepted errors (Appendix~\ref{app:classes}).

The exchanged sets are riskier than the shared core. Pooled error rates were \lbExchangeCommonRisk\% for epochs accepted by both FixedScale and Linear-BCE, \lbExchangeFixedOnlyRisk\% for FixedScale alone, and \lbExchangeBceOnlyRisk\% for Linear-BCE alone. This is consistent with the boundary-focused objective of Section~\ref{sec:refinement}, although set exchange does not establish proximity to a score threshold. The risk target concerns the complete accepted set, not either exchanged subset.

\section{Discussion and Conclusion}
COVER separates learning primary-error information from refining how much a fixed classifier can accept. Auxiliary-informed learning recovers correct predictions missed by confidence alone, and fixed-scale refinement improves the locked operating point over BCE initialization in nested development. Similar matched-budget rankings and riskier exchanged sets are consistent with the boundary-focused objective, without establishing a uniform ranking advantage.

These findings concern a single 20-person cohort, one primary architecture, and one protocol seed. FixedScale was specified after a prespecified multi-condition campaign that did not meet its success gate, and the Linear-BCE anchor was chosen after the prespecified analysis. Nested development reuses frozen component outputs, limiting independence at the component level. CERT supports empirical rule selection, not a per-subject risk guarantee (Appendix~\ref{app:bound}).

A next evaluation should freeze the complete scoring and selection pipeline before applying it to previously unexposed subjects. A downstream model is also needed to measure error resolution and end-to-end computational savings, including auxiliary and scoring costs. Within the studied cohort, COVER provides an auditable acceptance mechanism that retains useful primary predictions and reserves unresolved epochs for a later model, a first step toward allocating cascade computation by first-stage reliability.

\section*{Reproducibility Statement}
Code, experimental protocols, and frozen-result reproduction are available at \url{https://github.com/Kevin11Kaikai/COVER-Selective-Sleep-Staging}. The supplementary material includes the manuscript sources, which compile independently with every referenced table and figure, and an archive of the saved scores and numeric inputs underlying the reported results, including the exploratory analyses removed from the appendix. It does not include EEG data, model weights, or all decision-replay inputs; Sleep-EDF-20 is publicly available. The original EDF-to-NPZ conversion script is not available, and Appendix~\ref{app:data} lists the preprocessing choices that are therefore undocumented; compilation reproducibility is thus distinct from end-to-end experimental reproduction.

\section*{Ethics Statement}
This study uses public sleep data and does not evaluate clinical deployment or human-review decisions. Subject and stage differences constrain clinical interpretation; empirical accepted risk does not establish clinical safety.

\section*{AI Use Statement}
LLM-based assistants supported research discussion, literature retrieval, drafting, implementation, and figure preparation. All reported numbers are computed from saved experimental outputs. The human authors remain responsible for all claims and for the submission.
\begingroup
\raggedright
\bibliography{references}
\endgroup
\bibliographystyle{iclr2027_conference}
\clearpage
\appendix
\section{Implementation and Provenance}\label{app:implementation}
\suppressfloats[t]
COVER is the framework and FixedScale its main implementation. All numeric inputs are saved results; this revision performs no fitting or statistical recomputation. The supporting archive preserves removed analyses, detailed feature definitions, and source fingerprints.

\subsection{Processed cohort and components}\label{app:data}
The processed cohort contains 39 NPZ files and 42,308 epochs with unique dataset/subject/record/epoch keys. Stored codes 0--4 denote Wake, N1, N2, N3, and REM. Each file stores Fpz--Cz at 100\,Hz as $n\times3000\times1$; epoch identity is its zero-based row. No subject, stage, or epoch is removed after this inventory is fixed. Loading only reshapes the arrays and casts them to float32, without additional fitted normalization or filtering. The RF uses the same epochs; training subjects alone determine feature transformations, class weights, and augmentation.

\paragraph{Unrecovered upstream processing.}
The saved inputs and component receipts identify the executed loader and extractor, but the preceding EDF-to-NPZ conversion has not been recovered. Annotation mapping, movement/unscored handling, wake trimming, and pre-storage filtering or scaling remain unknown. Saved decisions support replay from processed inputs, not a complete raw-EDF-to-result recipe. The record inventory is archived in the supporting material.

\begin{table}[htbp]
\caption{Experimental settings, combining component and evaluation configuration. Every method shares primary predictions and the risk target.}
\label{tab:settings}\label{tab:component_config}
\centering\small
\begin{tabular}{@{}p{.24\linewidth}p{.70\linewidth}@{}}
\toprule
Item & Setting \\
\midrule
Cohort and input & Sleep-EDF-20; 20 subjects, 39 recordings, 42,308 epochs. Fpz--Cz at 100\,Hz, 30-s epochs; five stages; no other channels or neighboring epochs. \\
Outer roles & 20 folds, each with 15 FIT / 4 CERT / 1 OUTER subjects. \\
Component fits & 15 component OOF fits and one final fit per outer fold; roles in Table~\ref{tab:component_roles}. \\
Primary structure & EEG Conformer; embedding 40; depth 6; $1\times3000$ samples. \\
Primary optimization & Adam; learning rate $10^{-3}$; batch 64; cosine schedule; at most 100 epochs; patience 20. \\
Checkpoint & Mean accuracy on two selection subjects; earliest epoch on ties. Class weights use component-training subjects only. \\
Auxiliary & RF with 35 features, 100 trees, maximum depth 10, balanced class weights; train-only SMOTE ratio 0.3 precedes StandardScaler. \\
RF probabilities & Isotonic calibration on two disjoint FIT subjects. \\
Risk vector & 14 primary + 14 auxiliary + 9 disagreement features. \\
Risk cross-validation & Five folds, each with 12 risk-training and 3 validation subjects. \\
FixedScale & $\kappa\in\{0.01,0.1,1,10\}$, then $\gamma\in\{0.01,0.1,1,10\}$; $T_0=0.2$; SLSQP starts $\beta_0,4\beta_0$; at most 400 iterations per attempt. \\
Acceptance & $\alpha=0.05$; 20 subject-equal quantile rules; minimum $m=50$ accepted epochs per CERT subject. \\
Nested development & Twenty contexts; each holds out 3 of 15 FIT subjects in five folds; four inner 9/3 splits within the remaining 12. FixedScale uses $\gamma=1$. \\
Matched counts & Linear-BCE's per-subject accepted counts; \lbMatchedBudget{} epochs in total. \\
Resampling and seed & 10,000 paired subject-bootstrap draws; seed 20260901. Component fits use deterministic per-fit seeds from the manifest. \\
\bottomrule
\end{tabular}
\end{table}
\begin{table}[htbp]
\caption{Component fits within each outer fold. CERT and OUTER subjects never fit, select, or calibrate components.}
\label{tab:component_roles}
\centering\small
\begin{tabular}{@{}p{.22\linewidth}p{.36\linewidth}p{.36\linewidth}@{}}
\toprule
 & Leave-one-subject-out fits (15 per outer fold) & Final fit (one per outer fold) \\
\midrule
Purpose & Outputs for one held-out FIT subject & Outputs for CERT and OUTER subjects \\
Subject pool & The 14 FIT subjects other than the target & 14 of 15 FIT subjects; one left unused \\
Roles & 10 training / 2 checkpoint selection / 2 RF calibration & Same 10/2/2 structure, fresh assignment \\
Scored subjects & The held-out FIT subject & All CERT and OUTER subjects \\
Label use & Target labels form $e$ after outputs are fixed & None at this stage; CERT labels are read for rule locking, OUTER labels only for evaluation \\
Unused subject & --- & Its risk-training rows come from its own leave-one-subject-out fit \\
\bottomrule
\end{tabular}

\end{table}

\subsection{Auxiliary and risk features}\label{app:rf35}
The RF extractor operates independently on each stored epoch. Table~\ref{tab:rf_features} lists its 35 features; full formulas, library settings, and exception handling are archived. These features fit no population statistics. Table~\ref{tab:feature_schema} defines the separate risk vector. Ground truth, subject identity, neighboring epochs, raw RF features, and target normalization statistics never enter that vector.
\begin{table}[htbp]
\caption{RF input features, distinct from the risk vector in Table~\ref{tab:feature_schema}. Total delta is slow plus fast delta; $\epsilon=10^{-10}$.}
\label{tab:rf_features}
\centering\small
\begin{tabular}{@{}p{.10\linewidth}p{.81\linewidth}@{}}
\toprule
Indices & Features in stored order \\
\midrule
1--11 & Standard deviation; interquartile range; skewness; excess kurtosis; sign-change rate; Hjorth mobility and complexity; mean; root mean square; sample entropy; normalized permutation entropy \\
12--18 & Absolute band powers: slow delta (0.4--1 Hz), fast delta (1--4), theta (4--8), alpha (8--12), sigma (12--16), beta (16--30); their sum \\
19--24 & The same six band powers divided by their sum plus $\epsilon$ \\
25--29 & Total delta/theta, delta/sigma, delta/beta, alpha/theta, theta/sigma, with $\epsilon$ added to each denominator \\
30--33 & Spectral centroid, spectral entropy, 95\% spectral edge, spectral slope \\
34--35 & Mean squared db4 wavelet detail coefficients at levels D4 and D5 \\
\bottomrule
\end{tabular}

\end{table}
\begin{table}[htbp]
\caption{Risk-vector blocks. Primary-only learning uses the first block; the heterogeneous vector uses all $d=37$ features.}
\label{tab:feature_schema}
\centering\small
\begin{tabular}{@{}p{.30\linewidth}rp{.57\linewidth}@{}}
\toprule
Feature block & Count & Contents \\
\midrule
Primary & 14 & $p_T$ (5); primary class indicators (5); MSP, runner-up, margin, normalized entropy (4) \\
Auxiliary & 14 & $p_F$ (5); RF class indicators (5); MSP, runner-up, margin, normalized entropy (4) \\
Disagreement & 9 & Forward, reverse and symmetric KL; Jensen--Shannon divergence; probability L1/L2 distances; label disagreement; $p_T(\hat y_F)$ and $p_F(\hat y_T)$ \\
\bottomrule
\end{tabular}

\end{table}

\subsection{Candidate models and score calibration}
The primary-only comparator searches the eight models in Table~\ref{tab:risk_config}. Linear-BCE and its refinements use the four logistic configurations on 37 features. Epoch weights normalized to mean one in the implementation yield Equation~\ref{eq:bce} after division by $N$.
\begin{table}[htbp]
\caption{Risk-model candidates. The primary-only search uses all eight; the linear group uses the logistic configurations with its shared calibration.}
\label{tab:risk_config}
\centering\small
\begin{tabular}{@{}p{.27\linewidth}p{.67\linewidth}@{}}
\toprule
Model & Settings \\
\midrule
Logistic regression (4) & L2; $C=\kappa\in\{0.01,0.1,1,10\}$; lbfgs; 2,000 iterations; tolerance $10^{-8}$; training-only StandardScaler. \\
Histogram boosting (4) & Log loss; 7/15 leaves $\times$ L2 penalties 0/1; learning rate 0.05; 200 iterations; minimum leaf size 100; 64 bins; no early stopping. \\
Subject weights & Inverse subject epoch counts, normalized to mean one; no risk-model SMOTE or class weights. \\
\bottomrule
\end{tabular}
\end{table}

Risk-level OOF fits exclude the target subjects. For each three-subject validation fold, a Platt map fits the other folds' raw OOF scores and errors. Concatenated calibrated validation scores select configurations by feasible coverage, then risk. BCE ties prefer smaller $C$; refinement retains that BCE choice and prefers larger $\gamma$ on ties. Primary-only ties prefer logistic regression, smaller $C$, then lexical configuration ID. An empty feasible rule set scores zero.

Validation subjects are absent as direct rows from the risk and Platt fits scoring them. However, the other OOF scores used to fit their Platt map can have risk-model ancestors trained on those validation subjects, and component outputs are shared. Thus this is conditional validation, not a fully nested component-training chain. CERT and OUTER remain excluded from scorer fitting and configuration selection.

The selected risk model is refitted on all FIT rows. Its final Platt map fits the pooled raw OOF scores, then acts on refitted scores. This calibration transfer can change available CERT operating points; CERT evaluates the final scorer's own scores. Candidate thresholds retain their calibrated validation-score origin rather than being recomputed in sample.

\paragraph{Shared linear calibration.}
BCE, SELE, FixedScale, and BCEAnchor standardize raw scores with calibration-training weighted mean $\mu$ and deviation $\sigma$. With $v=(u-\mu)/\sigma$, they fit
\begin{equation}
\min_{a\ge0,b}\ \sum_{s,i}w_{si}[\log(1+\exp(av_{si}+b))-e_{si}(av_{si}+b)]+\frac{a^2}{2N}.
\end{equation}
The decision score is the calibrated logit $av+b$, not its sigmoid; ordering is nondecreasing. A constant score uses zero slope and the subject-weighted error-prior logit. Fitting fails closed if an error class is absent. The primary-only pipeline instead clips error probabilities to $[10^{-6},1-10^{-6}]$, takes logits, and fits an unconstrained-slope L2 Platt map with $C=1$. Detailed optimizer tolerances are archived.

\subsection{Acceptance and matched ranking}\label{app:acceptance}
For quantiles $q\in\{1/20,\ldots,1\}$ of FIT calibrated validation scores, let $B(\tau)$ and $E(\tau)$ be subject-weighted mass below and equal to $\tau$. Choose the threshold spanning $q$ and store $\rho=(q-B)/E$ as an exact rational. Rule $\pi=(\tau,\rho)$ accepts
\begin{equation}
a_{si}=\ind\{\hat r_{si}<\tau\}+\ind\{\hat r_{si}=\tau\}\ind\{h_{si}<\rho\}.
\label{eq:acceptance}
\end{equation}
Equality uses stored float64 scores. The label-free SHA256 tie value binds protocol, outer fold, dataset, subject, recording, epoch, and method, excluding $q$ so candidates are nested. Linear methods share one hash namespace; exact strings are in the supporting protocol record.

CERT requires defined $\widehat R_{\rm SE}\le\alpha$ and at least 50 accepted epochs per subject. It maximizes realized subject coverage, then minimizes exact risk, then $q$. Infeasibility returns all-abstention; FIT selection does not impose the support requirement. Matched ranking sorts risk scores with a common SHA256 order of dataset/subject/record/epoch and retains the prescribed count per subject without outcome labels.

For reference, the coverage-weighted identity underlying Equation~\ref{eq:metrics} is
\begin{equation}
\RSE=\frac{\sum_{s:A_s>0}(A_s/n_s)(K_s/A_s)}{\sum_{s:A_s>0}A_s/n_s}.
\label{eq:rse_weighted}
\end{equation}
It is not an equally weighted mean of conditional subject risks.

\subsection{Baseline scoring and fusion}\label{app:scores}
MSP uses $1-\max_jp_T(j)$; forward-KL uses $D_{\rm KL}(p_T\Vert p_F)$; primary-only uses its selected calibrated error model. All scores increase toward rejection and retain the primary label.
\paragraph{DuoF.}\label{app:duof}
DuoF fits two nonnegative weights bounded by 20 to subject-weighted multiclass negative log-likelihood on FIT OOF outputs. Primary logits and log RF probabilities form fused probabilities. Risk is one minus the fused probability assigned to the original primary class, irrespective of the fused argmax. Weights are not cross-fitted before FIT quantiles are computed; thus DuoF and COVER differ in loss, capacity, and selection.

\subsection{Linear ranking and refinement}\label{app:refinement}
SELE shares the 37 features, training-only scaler, and subject weights, using the pairwise hinge surrogate of \citet{franc2019uncertainty}:
\begin{equation}
\min_\theta\ \sum_{(s,i),(s',j)}w_{si}w_{s'j}e_{si}[1+s_\theta(z_{s'j})-s_\theta(z_{si})]_+
+\frac{\eta}{2}\|\theta\|_2^2,\quad s_\theta(z)=z^\top\theta,
\label{eq:sele}
\end{equation}
where $\eta=1/(NC)$. Sorting computes the full weighted hinge loss and subgradient; a convex solver checks the optimality gap. Each candidate starts from BCE at its own $C$, and SELE selects $C$ independently of FixedScale.

FixedScale uses $\sigma_0\ge10^{-6}$, $\beta_0=\theta_{\rm BCE}/\sigma_0$, and $T_0=0.2$. BCEAnchor changes the penalty center from zero to $\beta_0$. SLSQP uses analytic gradients, starts $\beta_0$ and $4\beta_0$, and at most 400 iterations with two attempts per start. Each start uses the best feasible of 104 thresholds: 101 subject-weighted quantiles and three values below the smallest score. Risk is imposed as $C_{\rm soft}-\sum w_{si}e_{si}g_{si}/\alpha\ge0$, with tolerance $10^{-6}$. The best converged feasible solution is retained; raw scores restore the fixed BCE scale and intercept before calibration, and the training threshold is discarded. Saved diagnostics report feasible solutions for all 435 FixedScale fits; 869 of 871 SLSQP runs converged on their first attempt.

\section{Component Performance}\label{app:components}
All 42,308 epochs are scored once as OUTER epochs. Conformer labels are unchanged; RF labels are the argmax of calibrated probabilities. Pooled metrics use all epochs, while subject means compute each subject's metrics first. Class-macro metrics average the five classes; undefined precision or recall contributes zero. Table~\ref{tab:components} characterizes the underlying models; Appendix~\ref{app:classes} reports stage-specific acceptance.
\begin{table}[htbp]
\caption{Component aggregate metrics (\%). Pooled epochs use one confusion matrix; mean subject averages 20 subject summaries.}
\label{tab:components}
\centering\small
\begin{tabular}{llrrrr}
\toprule
Component & Aggregation & Accuracy & Macro P & Macro R & Macro-F1 \\
\midrule
EEG Conformer & Pooled epochs & 76.01 & 67.80 & 71.07 & 69.02 \\
EEG Conformer & Mean subject & 75.86 & 69.25 & 71.56 & 67.76 \\
RF (isotonic) & Pooled epochs & 77.52 & 68.23 & 67.77 & 67.35 \\
RF (isotonic) & Mean subject & 77.58 & 68.74 & 68.07 & 65.99 \\
\bottomrule
\end{tabular}

\end{table}
\section{Additional Comparisons and Robustness}\label{app:composition}
\subsection{Matching anchors and prespecified comparisons}
The main text uses Linear-BCE's own-rule counts, totaling \lbMatchedBudget{} accepted epochs. This anchor was chosen after the prespecified analysis to match the step-(i) scorer. The original pipeline, termed Original COVER, selected among 16 logistic and boosting configurations across primary-only and heterogeneous features. Its prespecified matched comparisons use its own counts, totaling 19,892 epochs (Table~\ref{tab:matched}). Earlier saved-output analysis reproduced the original counts and intervals before computing the Linear-BCE comparisons; neither matching anchor retrains models or changes own-rule outcomes.

At the newer budget, Original COVER incurs \lbOriginalMatchedErrors{} errors; its pooled-risk difference from Linear-BCE is \lbOriginalMinusLinearPoolCI{} pp. Table~\ref{tab:linear_pairs} gives the Linear-BCE comparisons used in the main text. It has fewer matched errors than MSP for \lbMspFewer/20 subjects. Against DuoF, one subject contributes \lbDuofSubjectFourteen{} of \lbDuofFewerTotal{} fewer errors; delete-one pooled-risk differences span \lbDuofDeleteRange{} pp.
\begin{table}[htbp]
\caption{Prespecified matched comparisons at 19,892 accepted epochs, with identical counts within every subject. COVER denotes Original COVER; $\Delta R$ is its pooled risk minus the comparator (pp).}
\label{tab:matched}
\centering\small
\begin{tabular}{lrrrrl}
\toprule
Score & Errors & Risk & Pass & $\Delta R$ & Descriptive 95\% interval \\
\midrule
COVER & 817 & 4.11 & 12/20 & -- & reference \\
MSP & 1,369 & 6.88 & 9/20 & -2.77 & [-5.11, -1.13] \\
DuoF & 946 & 4.76 & 14/20 & -0.65 & [-2.24, 0.36] \\
Primary-only & 1,133 & 5.70 & 10/20 & -1.59 & [-3.39, -0.34] \\
Forward-KL & 2,171 & 10.91 & 3/20 & -6.81 & [-9.64, -4.46] \\
\bottomrule
\end{tabular}

\end{table}

\begin{table}[htbp]
\caption{Linear-BCE comparisons at the main-text matching budget. Values are saved differences and descriptive 95\% intervals, in pp; directions follow the first column.}
\label{tab:linear_pairs}
\centering\small
\begin{tabular}{@{}llr@{}}
\toprule
Comparison & Endpoint & Difference [95\% interval] \\
\midrule
Linear-BCE -- MSP & Own-rule $C$ & \lbMspCovCI{} \\
Linear-BCE -- MSP & Own-rule $R_{\rm SE}$ & \lbMspRiskCI{} \\
Linear-BCE -- MSP & Matched $R_{\rm pool}$ & \lbMspPoolCI{} \\
Linear-BCE -- Primary-only & Own-rule $C$ & \lbPrimaryCovCI{} \\
Linear-BCE -- Primary-only & Matched $R_{\rm pool}$ & \lbPrimaryPoolCI{} \\
Linear-BCE -- DuoF & Own-rule $C$ & \lbDuofCovCI{} \\
Linear-BCE -- DuoF & Matched $R_{\rm pool}$ & \lbDuofPoolCI{} \\
FixedScale -- Linear-BCE & Matched $R_{\rm pool}$ & \lbFixedMinusLinearPoolCI{} \\
Original COVER -- Linear-BCE & Own-rule $C$ & \lbOriginalMinusLinearCovCI{} \\
Original COVER -- Linear-BCE & Matched $R_{\rm pool}$ & \lbOriginalMinusLinearPoolCI{} \\
\bottomrule
\end{tabular}
\end{table}

\begin{figure}[htbp]
\centering\includegraphics[width=\linewidth]{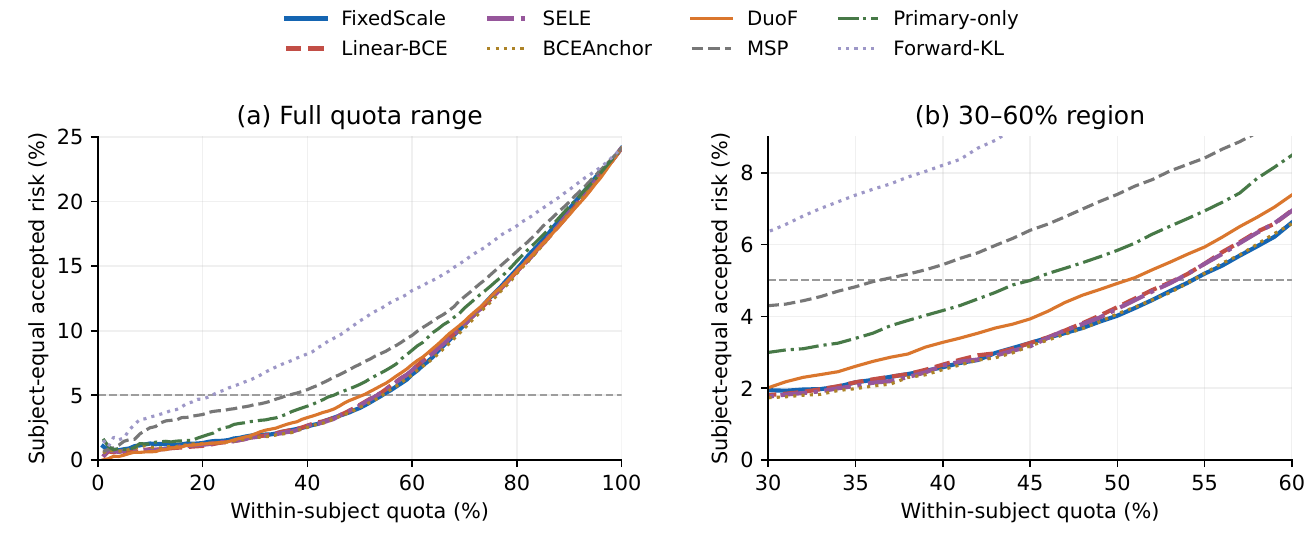}
\caption{Ranking diagnostics for the eight main-text scorers from saved scores. Left: full within-subject quota range; right: the 30--60\% region. Risks are subject-equal; the dashed line marks 5\%. OUTER outcomes diagnose ranking and never select deployed rules. Own-rule points need not lie on these curves because per-subject acceptance allocations differ.}
\label{fig:quota}
\end{figure}
Within the 30--60\% region, the saved Linear-BCE, FixedScale, and SELE curves differ by at most \lbCurveLinearSpread{} pp. Similar quota performance does not establish a uniform ranking advantage; locked rules can allocate acceptance differently across subjects.

\subsection{Nested development}\label{app:upgrade}
Each of 20 contexts forms five folds holding out three FIT subjects from 12 risk-training subjects. Four inner 9/3 splits select configurations and rules within each parent 12-subject set. Scalers, BCE/SELE parameters, and score scales are refitted inside the risk layer; frozen component OOF outputs are reused. Cross-calibrated inner scores select BCE $C$, whose initialization supplies FixedScale-1 and the Anchor variants. FixedScale-1 and BCEAnchor-1 fix strength one; BCEAnchor-selected chooses among four strengths. SELE selects its own $C$. Exact picked-quantile priority, then risk, then stronger Anchor regularization or smaller SELE $C$ resolve configuration choices.

Inner scores fix threshold, tie fraction, and quantile before the parent's held-out subjects are processed. Final risk fitting uses the 12 parent subjects; the final Platt map uses their held-out raw scores. The saved inner rule is applied unchanged to the three held-out subjects, whose labels affect evaluation only.

For each subject, average repeated acceptance and error masses: $\bar c_s=\operatorname{mean}_j(A_{sj}/n_{sj})$ and $\bar e_s=\operatorname{mean}_j(K_{sj}/n_{sj})$. Report $C=S^{-1}\sum_s\bar c_s$ and $\RSE=\sum_s\bar e_s/\sum_s\bar c_s$. The 300 records therefore contribute 20 unique subject units. FixedScale-1 has greater coverage than BCE for \rvNestedFixedBceCovHigher/20 subjects. This remains descriptive development within the same cohort.
\begin{table}[htbp]
\caption{Nested development within FIT. BCE denotes Linear-BCE. Locked columns apply frozen inner rules; quota risk fixes 50\% acceptance per subject. Values are percentages; $\Delta C$ is row minus FixedScale-1 in pp. Zero counts no-acceptance records among 300; aggregation uses 20 subjects.}
\label{tab:nested}
\centering\footnotesize
\begin{tabular}{@{}lrrrrl@{}}
\toprule
Scorer & Locked $C$ & Locked $R_{\rm SE}$ & Quota $R_{\rm SE}$ & Zero & \shortstack{$\Delta C$ vs. FixedScale-1\\{}[95\% CI], pp} \\
\midrule
\textbf{FixedScale-1} & 50.21 & 4.45 & 4.69 & 3 & -- \\
BCE & 48.56 & 4.84 & 5.03 & 9 & -1.65 [-2.52, -0.47] \\
SELE & 49.62 & 4.96 & 5.07 & 12 & -0.60 [-1.45, +0.50] \\
BCEAnchor-1 & 49.55 & 4.64 & 4.88 & 6 & -0.66 [-1.27, +0.15] \\
BCEAnchor-selected & 49.83 & 4.73 & 4.92 & 6 & -0.39 [-1.05, +0.54] \\
\bottomrule
\end{tabular}

\end{table}

\subsection{Paired uncertainty and sensitivity}\label{app:revision}
Positive coverage differences and negative risk differences favor the first named method in Table~\ref{tab:revisionpairs}. In the historical own-rule comparison, FixedScale coverage is higher than DuoF for \rvHistFixedDuofCovHigher/20 subjects and higher than SELE for \rvHistFixedSeleCovHigher/20; the SELE coverage interval includes zero. The sensitivity table retains every subject in the primary estimate and reports all single-subject deletions separately.
\begin{table}[htbp]
\caption{Core paired differences (pp), with descriptive 95\% intervals. BCE denotes Linear-BCE. Locked and fixed-quota outcomes are distinct evaluations; all rows are copied from the saved comparison table.}
\label{tab:revisionpairs}
\centering\footnotesize
\begin{tabular}{@{}lllrrr@{}}
\toprule
Scope / comparison & Rule & Metric & Difference & Lower & Upper \\
\midrule
Nested: FixedScale-1 -- BCE & Locked & $C$ & +1.65 & +0.47 & +2.52 \\
Nested: FixedScale-1 -- BCE & Locked & $R_{\rm SE}$ & -0.39 & -1.41 & +0.23 \\
Nested: FixedScale-1 -- BCE & 50\%-quota & $R_{\rm SE}$ & -0.34 & -1.30 & +0.24 \\
Nested: FixedScale-1 -- SELE & Locked & $C$ & +0.60 & -0.50 & +1.45 \\
Nested: FixedScale-1 -- SELE & Locked & $R_{\rm SE}$ & -0.51 & -1.41 & +0.05 \\
Nested: FixedScale-1 -- SELE & 50\%-quota & $R_{\rm SE}$ & -0.37 & -1.45 & +0.27 \\
Nested: BCEAnchor-1 -- FixedScale-1 & Locked & $C$ & -0.66 & -1.27 & +0.15 \\
Nested: BCEAnchor-1 -- FixedScale-1 & Locked & $R_{\rm SE}$ & +0.19 & -0.26 & +0.88 \\
Nested: BCEAnchor-1 -- FixedScale-1 & 50\%-quota & $R_{\rm SE}$ & +0.19 & -0.27 & +0.93 \\
Nested: BCEAnchor-selected -- FixedScale-1 & Locked & $C$ & -0.39 & -1.05 & +0.54 \\
Nested: BCEAnchor-selected -- FixedScale-1 & Locked & $R_{\rm SE}$ & +0.28 & -0.23 & +1.07 \\
Nested: BCEAnchor-selected -- FixedScale-1 & 50\%-quota & $R_{\rm SE}$ & +0.23 & -0.26 & +1.03 \\
OUTER: FixedScale -- SELE & Locked & $C$ & +1.05 & -3.71 & +6.58 \\
OUTER: FixedScale -- SELE & Locked & $R_{\rm SE}$ & +0.21 & -0.45 & +0.86 \\
OUTER: FixedScale -- DuoF & Locked & $C$ & +6.78 & +1.54 & +12.76 \\
OUTER: FixedScale -- DuoF & Locked & $R_{\rm SE}$ & +0.24 & -0.89 & +1.42 \\
OUTER: FixedScale -- BCE & Locked & $C$ & +1.61 & -3.67 & +7.30 \\
OUTER: FixedScale -- BCE & Locked & $R_{\rm SE}$ & +0.38 & -0.29 & +1.03 \\
\bottomrule
\end{tabular}

\end{table}
\begin{table}[htbp]
\caption{Delete-one sensitivity of nested differences (pp). Min./max. span all 20 deletions; the identified column is a sensitivity analysis, not a subject exclusion from primary results.}
\label{tab:revisionsensitivity}
\centering\footnotesize
\begin{tabular}{@{}llrrr@{}}
\toprule
Comparison & Metric & Delete-one min. & Max. & Without S11 \\
\midrule
FixedScale-1 -- BCE & $C$ & +1.50 & +2.12 & +2.12 \\
FixedScale-1 -- BCE & Locked $R_{\rm SE}$ & -0.47 & +0.08 & +0.08 \\
FixedScale-1 -- BCE & Quota $R_{\rm SE}$ & -0.43 & +0.09 & +0.09 \\
FixedScale-1 -- SELE & $C$ & +0.44 & +1.01 & +1.01 \\
FixedScale-1 -- SELE & Locked $R_{\rm SE}$ & -0.58 & -0.10 & -0.10 \\
FixedScale-1 -- SELE & Quota $R_{\rm SE}$ & -0.47 & +0.12 & +0.12 \\
BCEAnchor-1 -- FixedScale-1 & $C$ & -0.98 & -0.55 & -0.98 \\
BCEAnchor-1 -- FixedScale-1 & Locked $R_{\rm SE}$ & -0.12 & +0.26 & -0.12 \\
BCEAnchor-1 -- FixedScale-1 & Quota $R_{\rm SE}$ & -0.14 & +0.27 & -0.14 \\
BCEAnchor-selected -- FixedScale-1 & $C$ & -0.75 & -0.27 & -0.75 \\
BCEAnchor-selected -- FixedScale-1 & Locked $R_{\rm SE}$ & -0.08 & +0.36 & -0.08 \\
BCEAnchor-selected -- FixedScale-1 & Quota $R_{\rm SE}$ & -0.12 & +0.32 & -0.12 \\
\bottomrule
\end{tabular}

\end{table}

\paragraph{Resampling.}\label{app:resampling}
Each draw samples 20 subject indices with replacement and carries all their acceptance and error masses; paired comparisons use the same draw. Ratios follow Equation~\ref{eq:metrics}. Percentile intervals use linear interpolation at 2.5 and 97.5\%, with 10,000 PCG64 draws and seed 20260901. Undefined-risk draws are excluded only for that statistic; zero coverage remains in coverage means. Overlapping folds and cohort reuse make these intervals descriptive.

\paragraph{Empirical selection and finite-sample bounds.}\label{app:bound}
Four CERT subjects do not establish per-person risk guarantees. For the conservative construction $h=\sqrt{\log(2K/\delta)/(2M)}$, $U=\min\{1,(\bar Q+h)/(\bar C-h)\}$ when $\bar C>h$, $M=4$, $K=20$, and $\delta=0.05$ give $h\approx0.914$. Even $\bar Q=0,\bar C=1$ yields a capped bound of one. This specific construction is uninformative at 5\%; it is not a claim about all risk-control methods. COVER uses empirical CERT selection.

Exploratory controls (scorer search, selection criteria, alternative nested branches) are archived in the supporting material.

\section{Stage-wise Retention and Accepted Sets}
\suppressfloats[t]\label{app:classes}
True-stage groups distinguish retained correctness from predicted-class precision. For true class $c$, let $N_c$ be available epochs, $B_c$ primary-correct epochs, $A_c$ accepted epochs, $Q_c$ correctly accepted epochs, and $K_c=A_c-Q_c$ errors. Correct and incorrect rejections number $B_c-Q_c$ and $N_c-B_c-K_c$. Define
\begin{equation}
\begin{aligned}
\mathrm{AcceptedAccuracy}_c&=Q_c/A_c,\qquad
\mathrm{CorrectRetention}_c=Q_c/B_c,\\
\mathrm{CorrectAutomation}_c&=Q_c/N_c.
\end{aligned}
\label{eq:class}
\end{equation}
These are pooled true-stage diagnostics, distinct from subject-equal risk and precision among predictions labeled $c$. Zero denominators are undefined. Tables~\ref{tab:fixedclasses} and~\ref{tab:exchanges} use FixedScale's own frozen rules and identity-aligned primary outcomes.
\begin{table}[htbp]
\caption{FixedScale own-rule stage counts and rates. $N,B,A,Q,K$ use the definitions above; the last three columns are percentages.}
\label{tab:fixedclasses}
\centering\small
\begin{tabular}{@{}lrrrrrrrr@{}}
\toprule
True stage & $N$ & $B$ & $A$ & $Q$ & $K$ & $A/N$ & $Q/A$ & $Q/B$ \\
\midrule
Wake & 8,285 & 6,727 & 4,827 & 4,786 & 41 & 58.26 & 99.15 & 71.15 \\
N1 & 2,804 & 831 & 288 & 19 & 269 & 10.27 & 6.60 & 2.29 \\
N2 & 17,799 & 13,437 & 10,899 & 10,594 & 305 & 61.23 & 97.20 & 78.84 \\
N3 & 5,703 & 5,320 & 3,690 & 3,579 & 111 & 64.70 & 96.99 & 67.27 \\
REM & 7,717 & 5,844 & 1,751 & 1,495 & 256 & 22.69 & 85.38 & 25.58 \\
\bottomrule
\end{tabular}

\end{table}
\begin{table}[htbp]
\caption{FixedScale versus Linear-BCE accepted-set exchange, grouped by true stage. Common means accepted by both.}
\label{tab:exchanges}
\centering\small
\begin{tabular}{@{}lrrrrrr@{}}
\toprule
& \multicolumn{2}{c}{FixedScale only} & \multicolumn{2}{c}{BCE only} & \multicolumn{2}{c}{Common} \\
True stage & Epochs & Errors & Epochs & Errors & Epochs & Errors \\
\midrule
Wake & 653 & 12 & 287 & 14 & 4,174 & 29 \\
N1 & 85 & 76 & 41 & 34 & 203 & 193 \\
N2 & 1,065 & 62 & 678 & 50 & 9,834 & 243 \\
N3 & 232 & 25 & 180 & 17 & 3,458 & 86 \\
REM & 286 & 69 & 319 & 14 & 1,465 & 187 \\
\bottomrule
\end{tabular}

\end{table}

The shared set contains \rvExchangeCommonAccepted{} epochs and \rvExchangeCommonErrors{} errors; FixedScale-only and BCE-only sets contain \rvExchangeFixedOnlyAccepted{} and \rvExchangeBceOnlyAccepted{} epochs, with \rvExchangeFixedOnlyErrors{} and \rvExchangeBceOnlyErrors{} errors. The exchanged sets have higher pooled error rates than the shared core, consistent with the boundary-focused objective. Their membership alone does not establish distance from a score threshold. The risk constraint concerns the entire accepted set, not either exchanged subset.

\end{document}